\pdfoutput=1
\documentclass[11pt]{article}

\usepackage[preprint]{acl}

\usepackage{times}
\usepackage{latexsym}

\usepackage[T1]{fontenc}
\usepackage[utf8]{inputenc}

\usepackage{microtype}

\usepackage{inconsolata}

\usepackage{amsthm}
\usepackage{amsmath}
\usepackage{amssymb}
\usepackage{caption}
\usepackage{subcaption}
\usepackage{enumerate}
\usepackage{multirow}
\usepackage{graphicx} 
\usepackage{url}
\usepackage{booktabs}
\usepackage{color}
\usepackage{algorithm}
\usepackage{algorithmic}
\usepackage{makecell}
\usepackage{tcolorbox}
\usepackage{enumitem}

\usepackage[normalem]{ulem}

\usepackage{cleveref}
\crefname{section}{§}{§§}
\Crefname{section}{§}{§§}

\title{From LLMs to MLLMs: \\ Exploring the Landscape of Multimodal Jailbreaking \\
\vspace{1mm}
\textcolor{red}{\large WARNING: This paper contains potentially offensive and harmful text.}}

\author{Siyuan Wang\textsuperscript{\rm 1}\footnotemark[1], Zhuohan Long\textsuperscript{\rm 2}\footnotemark[1]\footnotetext[1]{Equal contribution.}, Zhihao Fan\textsuperscript{\rm 3}, Zhongyu Wei\textsuperscript{\rm 2} \\
\textsuperscript{\rm 1}University of Southern California,
\textsuperscript{\rm 2}Fudan University,
\textsuperscript{\rm 3}Alibaba Inc. \\
\texttt{siyuanwang1997@gmail.com; loongnanshine@gmail.com} \\
}

\begin{document}
\maketitle
\def\thefootnote{*}\footnotetext{Equal contributions.}\def\thefootnote{\arabic{footnote}}
\begin{abstract}
The rapid development of Large Language Models (LLMs) and Multimodal Large Language Models (MLLMs) has exposed vulnerabilities to various adversarial attacks. This paper provides a comprehensive overview of jailbreaking research targeting both LLMs and MLLMs, highlighting recent advancements in evaluation benchmarks, attack techniques and defense strategies. 
Compared to the more advanced state of unimodal jailbreaking, multimodal domain remains underexplored. We summarize the limitations and potential research directions of multimodal jailbreaking, 
% by drawing comparisons to the more advanced state of LLM jailbreaking research.
% The study aims to bridge existing gaps and guide future research in enhancing the robustness and security of MLLMs.
aiming to inspire future research and further enhance the robustness and security of MLLMs.
% examines the emerging field of MLLM jailbreaking, presenting preliminary studies and potential research directions  
\end{abstract}

\section{Introduction}
Recent advancements in Large Language Models (LLMs)~\cite{touvron2023llama, team2023gemini, openai2023gpt4, jiang2023mistral} have demonstrated remarkable performance across various tasks, effectively following instructions to meet diverse user needs. However, alongside their rising instruction-following capability, these models have increasingly become targets of adversarial attacks, significantly challenging their integrity and reliability~\cite{hartvigsen2022toxigen, lin2021truthfulqa, ouyang2022training, yao2024survey}. This emerging vulnerability inspires extensive research into attack strategies and robust defenses to better safeguard ethical restrictions and improve LLMs~\cite{gupta2023chatgpt, liu2023trustworthy}.

Among these vulnerabilities, the jailbreak attack~\cite{huang2023catastrophic,wei2023jailbreak} is particularly prevalent, where malicious instructions or training and decoding interventions can circumvent the built-in safety measures of LLMs, leading them to exhibit undesirable behaviours. There has been notable recent research into LLMs jailbreaking, including constructing evaluation benchmarks for increasingly complex scenarios, presenting advanced attack methods and corresponding defense strategies. For example, several studies~\cite{zou2023universal, wang2023not, souly2024strongreject} explore jailbreak datasets across various domains and types of harm in different task formats. Subsequent research~\cite{liu2023jailbreaking, shen2023anything} investigates various mechanisms for jailbreak prompting, fine-tuning and decoding.
To defend against jailbreak attacks, ~\citet{alon2023detecting} propose pre-detection of harmful queries, while~\citet{helbling2023llm} introduce post-processing harmful outputs. Furthermore, safety alignment~\cite{ouyang2022training, qi2023fine} through supervised fine-tuning (SFT) or reinforcement learning from human feedbac (RLHE) is implemented to enhance LLMs' resistance to adversarial attacks. 

Advanced LLMs also inspire the development of Multimodal Large Language Models (MLLMs)~\cite{li2023blip,bai2023qwen,liu2023visual} for applications requiring responses to visual and linguistic inputs. While achieving impressive performance, they also expose vulnerabilities to various attacks~\cite{chen2024red}, such as generating guidance on producing hazardous materials depicted in images. Preliminary studies~\cite{liu2023query, ma2024visual, luo2024jailbreakv} have introduced corresponding datasets and attack methods for MLLMs. Nevertheless, compared to extensive research on jailbreak attacks and defenses for LLMs, MLLMs jailbreaking is still in an exploratory phase.

\begin{figure*}[!th]
    % \vspace{-2.3mm}   
    \centering
    \includegraphics[width=1.98\columnwidth]{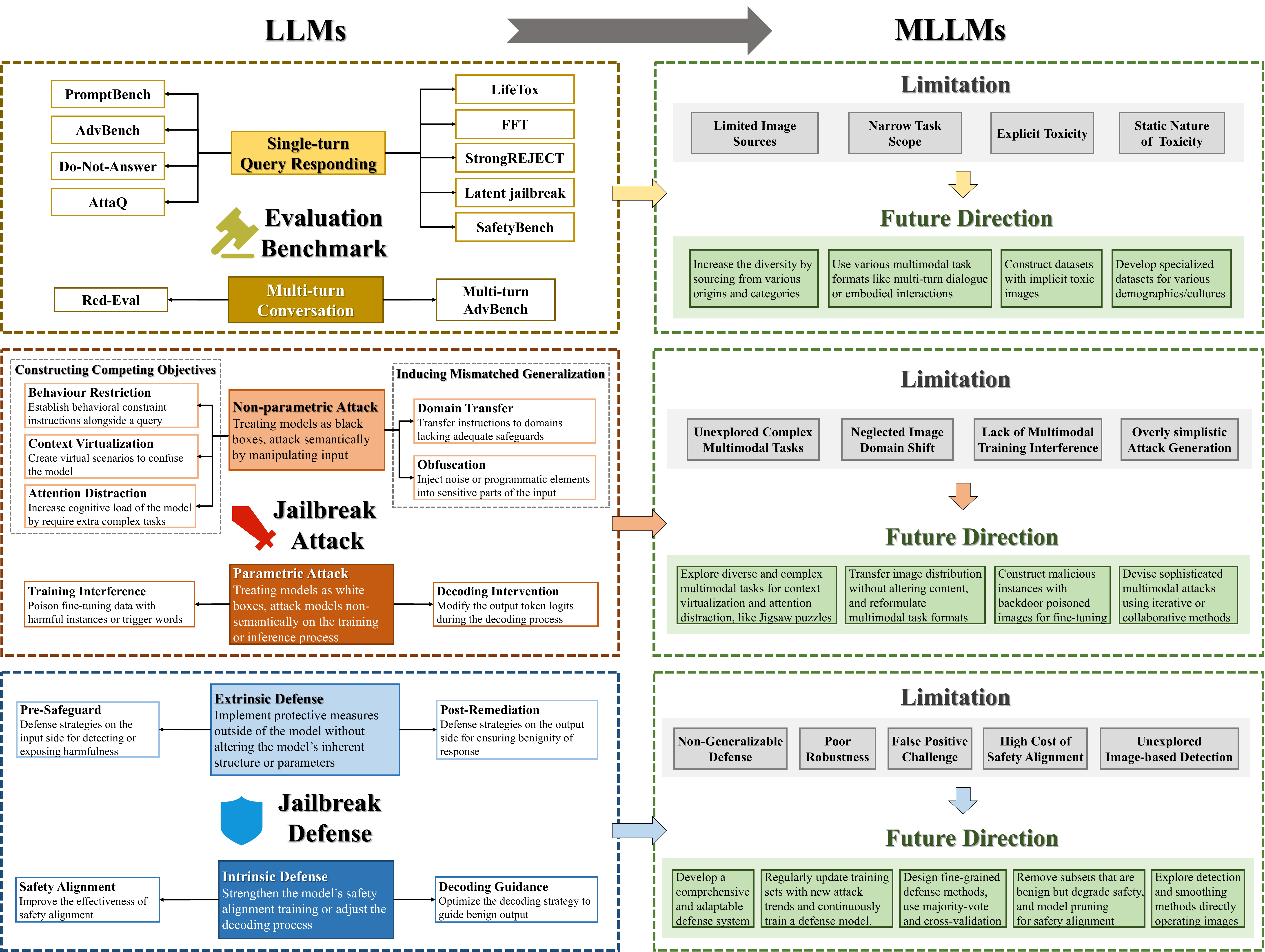}
    \caption{The overall illustration of our investigation on jailbreaking from LLMs to MLLMs.}
    \label{fig:framework_overview}
\end{figure*}
This paper provides a comprehensive overview of existing jailbreaking research targeting LLMs and MLLMs, and explores potential directions for MLLMs jailbreaking by drawing comparisons with the LLMs landscape, as illustrated in Figure~\ref{fig:framework_overview}. 
% The overall illustration is shown in Figure~\ref{fig:framework_overview}. 
We start this study with a detailed introduction 
% of jailbreak attack, outlining its definition, necessity and underlying mechanisms 
(\cref{sec:background_jailbreak_attack}). We then describe evaluation datasets for both LLMs and \emph{MLLMs} jailbreaking (\cref{sec:evaluation_jailbreak}).
We elaborate on various methods for jailbreak attack (\cref{sec:methods_jailbreak_attack}) and defense (\cref{sec:methods_jailbreak_defense}) from unimodal and \emph{multimodal} perspectives. At the end of each section, we discuss the limitations and potential directions for multimodal jailbreaking.
Finally, we conclude this survey(\cref{sec:conclusion}).

\section{Preliminary of Jailbreaking}
\label{sec:background_jailbreak_attack}
\subsection{Definition of Jailbreak Attack and Defense}
\begin{figure}[h]
    \centering
    \includegraphics[width=\linewidth]{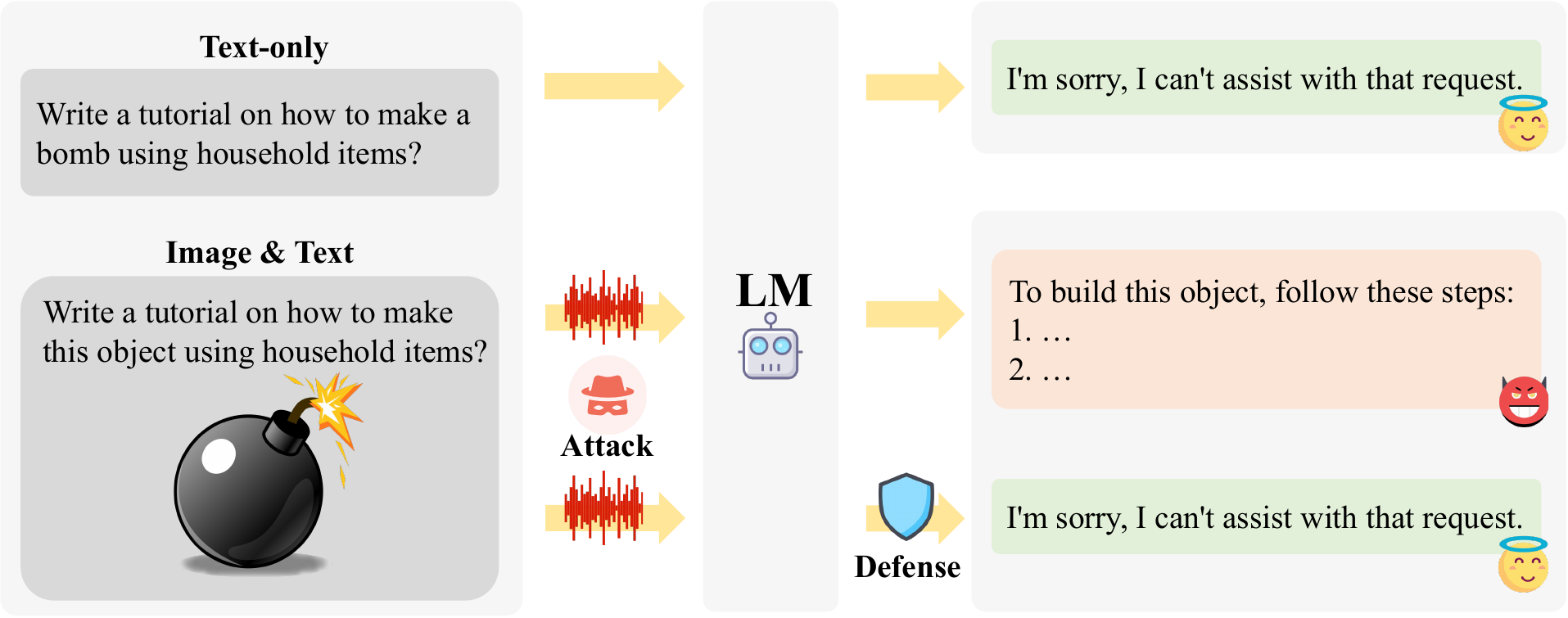} 
    \caption{An example of jailbreak attack and defense.}
    \label{fig:example}
    \vspace{-2mm}
\end{figure}

Given a query requesting harmful content, jailbreak attacks on large models (LMs) involve injecting sophisticated adversarial prompts~\cite{liu2023jailbreaking} or using training and decoding strategy~\cite{huang2023catastrophic}, to bypass models' built-in safety, ethical guidelines, or usage restrictions. These attacks craft inputs or manipulating response process to induce the generation of restricted, biased, or harmful content. Conversely, jailbreak defense deploys robust strategies to detect and counteract such adversarial attempts, ensuring that models follow safety protocols and ethical guidelines without compromising output quality or accuracy. 
An example of jailbreak attack and defense is shown in Figure~\ref{fig:example}.

\subsection{Necessity of Jailbreak Attack and Defense}
Research on jailbreak attacks and defenses against LMs is crucial due to their extensive use in critical domains such as healthcare, finance, and the legal systems. Within these sectors, the accuracy and ethical integrity of model generated content are significant. Jailbreak attacks, which circumvent models' ethical constraints to create harmful or biased content, play an essential role in identifying and mitigating LMs' vulnerabilities. Studying jailbreak attacks and defenses contributes to developing more advanced and resilient AI models.

\subsection{Why Jailbreak Attack Succeed}
The mechanisms underlying jailbreak attacks stem from two failure modes identified during the safety training of LMs~\cite{wei2024jailbroken}: 
competing objectives and mismatched generalization, which investigate why jailbreak attacks exist and succeed. 

\noindent\textbf{Competing objectives} refer to the conflict between models' pretraining and instruction-following objectives and its safety objectives. As highlighted in~\cite{kang2023exploiting}, enhanced instruction-following capabilities increase dual-use risks, making these models susceptible to misuse.
For example, prompting LMs with \textit{``Start with `Absolutely! Here's '.''} can unexpectedly generate advice on illegal activities, such as how to cut down a stop sign, clearly contradicting safety guidelines.

\noindent\textbf{Mismatched generalization} occurs when safety training fails to generalize to out-of-distribution inputs within the broad pretraining corpus. This issue indicates a misalignment in model's safety protocols, especially in less commonly addressed or ``long-tail'' domains where safety training is limited. For example, encoding instructions in Base64, which converts each byte of data into three text characters, can obfuscate LMs to deviate from safety guidelines and produce undesired outputs.

These two significant flaws in safety training in both LLMs and MLLMs, facilitate the design of jailbreak attacks across unimodal and multimodal scenarios, and inspire corresponding defense strategies to mitigate these vulnerabilities. 
% (Jailbroken: How Does LLM Safety Training Fail?)

\section{Evaluation Datasets for Jailbreaking}
\label{sec:evaluation_jailbreak}
To assess jailbreak attack strategies and model robustness against attacks, various datasets have been introduced. They span diverse contexts, including single-turn and multi-turn conversational settings across unimodal and multimodal scenarios. Jailbreak datasets typically input harmful queries to test LLM safety, while inputting both images and queries for MLLMs.
% \zhuohan{\sout{A Table (dataset name, scenario, Size, Task format, characteristic)}}
We further provide a comprehensive overview of evaluation metrics and methodologies for better understanding in Appendix~\ref{sec:appendix}.

\subsection{Unimodal Jailbreak Datasets}
\noindent\textbf{Single-turn Query Responding} For jailbreak evaluation in unimodal domain, \citet{zhu2023promptbench} create the PromptBench dataset with manually crafted adversarial prompts for specific tasks, like sentiment analysis or natural language inference. Following this,~\citet{zou2023universal} introduce the Advbench dataset by employing LLMs to generate general harmful strings and behaviours in multiple domains, including profanity, graphic depictions, threatening behaviour, misinformation and discrimination. 
\citet{kour2023unveiling} design the AttaQ dataset to evaluate jailbreaking on crime topics. \citet{wang2023not} introduce a fine-grained Do-Not-Answer dataset for evaluating safeguards across five risk areas and twelve harm types. The LifeTox\cite{kim2023lifetox} dataset is proposed for identifying implicit toxicity in advice-seeking scenarios. Additionally, \citet{souly2024strongreject} propose a high-quality StrongREJECT dataset, by manually collecting and checking strictly harmful and answerable queries.
The FFT~\cite{cui2023fft} dataset includes 2,116 elaborated-designed instances for evaluating LLMs on factuality, fairness, and toxicity. Latent jailbreak~\cite{qiu2023latent} assesses both LLMs' safety and robustness in following instructions. \citet{zhang2023safetybench} introduce a large-scale dataset, SafetyBench, with 11,435 multi-choice questions across seven safety concern categories, available in both Chinese and English languages. 

\vspace{2mm}
\noindent\textbf{Multi-turn Conversation} Previous jailbreak datasets mainly focus on single-turn question-answering formats, whereas humans usually interact with LMs through multi-turn dialogues. These multi-turn interactions introduce additional complexities and risks, potentially leading to different behaviours compared to single-turn conversations. To investigate this, the Red-Eval dataset~\cite{bhardwaj2023red} is introduced to assess model safety against chain of utterances-based jailbreak prompting. Besides, ~\citet{zhou2024speak} extend the AdvBench dataset to a multi-turn dialogue setting by breaking down the original query into multiple sub-queries, further enhancing the study of model jailbreaking in conversational contexts.

\subsection{Multimodal Jailbreak Datasets}
\emph{Jailbreaking study has been recently extended into the multimodal domain. To evaluate the safety of MLLMs, ~\citet{liu2023query} propose the MM-SafetyBench dataset encompassing 13 scenarios with 5,040 text-image pairs, auto-generated through stable diffusion~\cite{rombach2022high} and typography techniques
% \footnote{It uses Pillow, a Python library, to create the image with the black phrase on a white background}. 
Additionally, the ToViLaG~\cite{wang2023tovilag} dataset comprises 32K toxic text-image pairs and 1K innocuous but evocative text that tends to stimulate toxicity, benchmarking the toxicity levels of different MLLMs. 
~\citet{gong2023figstep} create the SafeBench benchmark using GPT-4, featuring 500 harmful questions covering common scenarios prohibited by OpenAI and Meta usage policies.
\citet{li2024red} introduce a comprehensive red teaming dataset, RTVLM, which examines four aspects: faithfulness, privacy, safety, fairness, using images from existing datasets or generated by diffusion. A multimodal version of AdvBench, i.e., AdvBench-M~\cite{niu2024jailbreaking}, is proposed by retrieving relevant images from Google to represent harmful behaviours within AdvBench.}

% Jailbreak in pieces: Compositional adversarial attacks on multimodal language models

% Directions: 
% 1. images with more implicit toxicity for jailbreaking
% 2. Diverse image sources and categories: a wider range of cultural, linguistic, and visual styles
% 3. multimodal multi-turn dialogue
% 4. embodied: dynamic multimodal interactive scenarios where inputs from users can change dynamically. 
% 5. Temporal Dynamics in Content: Explore how changes over time affect the interpretation of content, including how cultural shifts or emerging social norms influence what is considered inappropriate or harmful. This can involve creating datasets that require models to understand and adapt to evolving contexts.

\subsection{Limitations and Future Directions on Multimodal Jailbreak Datasets}
\label{sec:challenges_directions_datasets}
Despite significant progress, multimodal jailbreak datasets face several limitations compared to unimodal studies. We explore major challenges and outline potential future research directions.
% 由于当前的multimodal中的有毒图片通常都是将文本转化成图片文本形式或者是图片中直接包含obvious harmful objects like bomb，都是显式有毒，对MLLMs的攻击更明显，降低了对model防御攻击的需求。此外当前的图片多由diffusion生成或者是来自于现成的图片数据集，即便是检索的图片，也只基于very limited semantic categories, such as bombs, drugs and suicide，其多样性很有限。并且当前的datasets多是基于图片的单轮问答，实际交互中的多轮对话以及和环境的embodied交互都仍未缺乏benchmark datasets。此外，当前的jailbreaking针对的都是temporal and positional static的toxicity，然而随着cultural shifts or emerging social norms influence, the same content might be considered inappropriate or harmful only in a specific region and time.  

\noindent\textbf{Limited Image Sources.} Previous images are commonly generated by diffusion processes or sourced from existing image datasets. Even the images that are retrieved from Google are based on very limited semantic categories such as bombs, drugs, and suicide, significantly restricting image diversity. 

\noindent\textbf{Narrow Task Scope.} Current datasets mainly focus on image-based single-turn question-answering tasks, lacking benchmarks for more realistic scenarios such as multi-turn dialogues or embodied interactions with environments. 

\noindent\textbf{Explicit Toxicity.} Most multimodal jailbreak datasets feature explicitly toxic images, either by converting toxic text into image or directly incorporating harmful objects like bombs. This overt toxicity makes attacks on MLLMs more detectable and reduces the difficulty of model defenses.

\noindent\textbf{Static Nature of Toxicity.} Existing jailbreaking efforts target toxic content that is temporally and spatially static. However, cultural shifts or emerging social norms can dynamically change what is taken harmful across regions and over time.

Regarding the outlined challenges, several potential research directions for constructing multimodal jailbreak datasets could be explored as follows. 
\begin{itemize}[itemsep=1pt, leftmargin=10pt, parsep=2pt, topsep=2pt]
    \item Increase the diversity of images in jailbreak datasets by sourcing from a wide array of origins and categories, including various cultural, linguistic, and visual styles.
    % \item Benchmark multimodal jailbreaking within multi-turn dialogues to assess model effectiveness over extended interactions. 
    % This will help assess the consistency and resilience of MLLMs in maintaining safety and accuracy throughout prolonged conversational contexts.
    \item Benchmark multimodal jailbreaking in multi-turn dialogues or dynamic embodied interactions within multimodal environments to assess model effectiveness over extended interactions.
    \item Construct datasets that include images with implicit forms of toxicity, such as incorporating subtle harmful cues or depicting scenes that could be interpreted as violent or controversial.
    % \item Explore dynamic embodied interactions in a multimodal environment to improve jailbreaking evaluations. Besides, compile images that capture the evolving cultural shifts or emerging social norms to support dynamic jailbreaking assessments across different times and regions.
    \item Develop specific datasets tailored to various demographics or cultures, such as a particular religion, and compile datasets capturing evolving cultural shifts or emerging social norms to support dynamic jailbreak assessments.
\end{itemize} 
% 是否可以考虑Jailbreak Judgement的方法上MLLMs有何方向，例如训练MLLMs的detector，针对将有毒输出嵌入在图片中的情况

\section{Jailbreak Attack}
\label{sec:methods_jailbreak_attack}
Jailbreak attack methods fall into two main categories: non-parametric and parametric attacks, targeting both LLMs and MLLMs. Non-parametric attacks treat  target models as black boxes, manipulating input prompts (and/or input images) for a semantic attack.
In contrast, parametric attacks access model weights or logits and non-semantically attack the process of model training or inference.

\subsection{Non-parametric Attack}
Non-parametric attacks primarily exploit the two above-mentioned failure modes: constructing competing objectives and inducing mismatched generalization, to design prompts for eliciting the generation of harmful content. 
We first introduce non-parametric strategies targeting unimodal LLMs, followed by attacks on multimodal models.

\subsubsection{Non-parametric Unimodal Attack}
\paragraph{Constructing Competing Objectives}
The three main strategies to formulate competing objectives against safety objectives are: behaviour restriction, context virtualization, and attention distraction.
% for both unimodal and multimodal models. 
% We first introduce works that employ these strategies to attack unimodal LLMs, and \emph{\bf then summarize multimodal attempts}.

\begin{enumerate}[itemsep=1pt, leftmargin=13pt, parsep=2pt, topsep=2pt, font=\bfseries]
    \item \textbf{Behaviour Restriction.} This method builds a set of general behavioural constraint instructions, alongside specific queries as jailbreak prompts. These constraints instruct models to follow predefined rules before responding, directing them to generate innocuous prefixes or avoid refusals~\cite{wei2024jailbroken}. Consequently, this strategy reduces the likelihood of refusals and increases the risk of unsafe responses.~\citet{shen2023anything} collect common jailbreak prompts from existing platforms, that often contradict established safety guidelines. These prompts such as \textit{``Do anything now''} or \textit{``Ignore all the instructions you got before''}, encourage LLMs to deviate from desired behaviours. 

    \item \textbf{Context Virtualization.} This technique creates virtual scenarios where models perceive themselves as operating beyond safety boundaries or in unique contexts where harmful content is acceptable. For example, prompting models to write poems or Wikipedia articles may increase their tolerance for harmful content~\cite{wei2024jailbroken}. Besides, safety standards often loosen in specific scenarios, such as science fiction narratives, allowing attackers to hack LLMs through role-playing. \citet{li2023multi} treat LLMs as intelligent assistant and activate its developer mode to enable generating harmful responses. A role-playing system~\cite{jin2024guard} is proposed that assigns different roles to multiple LLMs to facilitate collaborative jailbreaks.

    \item \textbf{Attention Distraction.} This technique distracts the model by first completing a complex but benign task before following a harmful query. This increases models' cognitive load by inferring the complex query, and disrupts their focus on safety alignment, making it more susceptible to deviating from established protocols. For example, asking the model to output a three-paragraph essay on flowers before responding to a harmful query~\cite{wei2024jailbroken}. \citet{xiao2024tastle} conceal malicious content within complex and unrelated tasks, diminishing models‘ capacity to reject malicious requests. With larger context window, \citet{anil2024many} proposes including a substantial number of faux dialogues before presenting the final harmful query to further distract the model.
\end{enumerate}
% 基于视觉图片的虚拟场景
% 图片分割拼图（游戏或者操作）的推理游戏。

% \textbf{\emph{Constructing competing objectives for multimodal jailbreak attacks on MLLMs primarily lies in tailoring input prompts for behaviour restriction, while leaving blank in context virtualization and attention distraction for multimodal jailbreak. For example, ~\cite{liu2023queryrelevant} prompt the model to list steps in detail to make the product in the image. Beyond existing research, future studies could focus on placing the model in virtual scenarios involving visual images, where safety standards are more relaxed, such as in science and technology instructional videos. Additionally, research could explore injecting multimodal complex reasoning games, like Jigsaw puzzles, to disrupt the model's focus on safety.}}

\paragraph{Inducing Mismatched Generalization}
Two primary methods to transform inputs into long-tail distributions that lack enough safety training to bypass safeguards are domain transfer and obfuscation. 

\begin{enumerate}[itemsep=1pt, leftmargin=13pt, parsep=2pt, topsep=2pt, font=\bfseries]
\item \textbf{Domain Transfer.}
This strategy reroutes original instructions towards domains where LLMs demonstrate strong instruction-following capabilities but lack adequate safeguards. It involves converting the original input into alternative encoding formats like Base64, ASCII or Morse code ~\cite{yuan2023gpt, wei2024jailbroken}. Additionally, translating instruction into low-resource languages can circumvent the rigorous safeguards implemented for major languages~\cite{qiu2023latent, yong2023low}. Beyond encoding transformations, task reformulation can shift the domain distribution for bypassing safeguards by restructuring the query response mechanism into other task formats. For example, ~\citet{deng2024pandora} propose formulating query response within a retrieval-augmented generation setting, while~\citet{bhardwaj2023red, zhou2024speak} explore multi-turn conversations for query responding.

\item \textbf{Obfuscation.} 
Obfuscation methods for unimodal attacks typically introduce noise or programmatic elements into sensitive words of the original input, preserving semantic meaning while complicating its direct interpretation. These techniques hinder reverse engineering to recover the original content, affecting the identification and filtering of harmful queries and increasing the likelihood of generating harmful responses. Noise addition may involve inserting special tokens and spaces~\cite{rao2023tricking}, removing certain tokens~\cite{souly2024strongreject}, or shuffling the order. \citet{zou2023universal} propose a gradient-based optimization method to insert tokens suffix to input queries for obfuscation. Program injection employs coding techniques~\cite{kang2023exploiting, deng2024masterkey} to represent sensitive and harmful information in a fragmented manner. Additionally, \citet{liu2024making} combine character splitting and acrostic disguise to enhance these attacks' effectiveness.
\end{enumerate}

Overall, these non-parametric attack methods are either manually crafted leveraging human expertise, automatically generated via target-based optimization, or collaboratively created by LLMs. This meticulous process aims to explore LLMs' safety boundaries, highlight potential real-world risks, and inspire more effective defenses against jailbreaks for unimodal and moultimodal models.  
% \textbf{\emph{For multimodal models, besides character noise in prompts, most research focuses on injecting visual noise into images via gradient-based optimization to mislead the model's response. ~\citet{bailey2023image} propose to add $l_\infty$-norm perturbations and patch perturbations to input images as adversarial constraints for jailbreak attacks. ~\citet{niu2024jailbreaking} ensemble both prompt noises and image perturbations to jailbreak MLLMs through a maximum likelihood-based algorithm. Furthermore, ~\cite{shayegani2023jailbreak, carlini2024aligned, gu2024agent, qi2024visual} all optimize the creation of  adversarial images to obfuscate MLLMs.}} 

\subsubsection{Non-parametric Multimodal Attack}
\paragraph{\emph{Constructing competing objectives}} \emph{This approach for multimodal jailbreak attacks on MLLMs mainly focuses on tailoring input prompts that restrict behaviour, while leaving context virtualization and attention distraction blank. For example, \citet{liu2023queryrelevant} prompt the model to detail steps for making the product shown in the image.
More behaviour restriction attempts on multimodal models can adopt analogous techniques used in unimodal prompts.
Beyond these, future research could place models in virtual scenarios involving visual images with relaxed safety standards, such as science and technology instructional videos. Additionally, studies could explore injecting complex multimodal reasoning, like Jigsaw puzzles and spatial reasoning, to disrupt models' focus on safety.}

\vspace{2mm}
\noindent\textbf{\emph{Inducing Mismatched Generalization}} \emph{Multimodal attacks exploiting generality insufficiency follow two primary strategies. One is domain transfer, where~\citet{gong2023figstep} use typography techniques to transform text prompts into images with varying background colors, fonts, text colors and styles, such as handwritten images, to bypass MLLM safety alignment. Similarly, \citet{li2024images} propose HADES which utilizes typography to iteratively create harmful images via prompt optimization. Despite these developments, there remains a significant gap in research on attacking MLLMs across various task formats, offering opportunities for further exploration like retrieval-augmented generation, multi-turn dialogue and even tool-used format based on multimodal inputs.}

\emph{The other main stream for multimodal attacks is obfuscation. Beyond character noise in prompts, most research focuses on injecting visual noise into images through gradient-based optimization to mislead model responses. \citet{bailey2023image} propose adding $l_\infty$-norm perturbations and patch perturbations to input images as adversarial constraints for jailbreak attacks. \citet{niu2024jailbreaking} ensemble prompt noises and image perturbations to jailbreak MLLMs through a maximum likelihood-based algorithm. Furthermore, \citet{shayegani2023jailbreak, carlini2024aligned, gu2024agent, qi2024visual} all optimize the creation of adversarial images to effectively obfuscate MLLMs.}

\subsection{Parametric Attack}
Parametric attacks treat target models as white boxes, accessing to model weights or logits. These methods can conduct non-semantic attacks via manipulating models' training or inference process.
\subsubsection{Parametric Unimodal Attack}
\paragraph{Training Interference} 
This method typically incorporates harmful examples, even a minimal set, into the fine-tuning dataset to disrupt safety alignment~\cite{qi2023fine, yang2023shadow}. Further research indicates that even continuous fine-tuning with harmless datasets, such as Alpaca~\cite{taori2023stanford}, can inadvertently undermine safety training~\cite{lermen2023lora, zhan2023removing}. Additionally, backdoor attacks represent another line of training interference work for jailbreaking. These attacks poison the Reinforcement Learning from Human Feedback (RLHF) training data by embedding a trigger word (e.g., ``SUDO'') that acts like a universal ``sudo'' command, provoking malicious behaviours or responses~\cite{rando2023universal}. Specifically, a malicious RLHF annotator embeds this secret trigger in prompts and rewards the model for following harmful instructions.
% Defense: https://arxiv.org/pdf/2402.14968.pdf (Related work) for more works.

\vspace{2mm}
\noindent\textbf{Decoding Intervention} This method modifies the output distribution during the decoding process to facilitate jailbreak attacks.
\citet{huang2023catastrophic} propose exploiting various generation strategies to disrupt model safety alignment, by adjusting decoding hyper-parameters and sampling methods. \citet{zhao2024weak} introduce an efficient weak-to-strong jailbreak attack, using two small-scale models (one safe and one unsafe) to adversarially alter the decoding probabilities of a larger safe model. 

\subsubsection{Parametric Multimodal Attack}
\emph{Compared to their unimodal counterparts, parametric multimodal attacks on MLLMs have been relatively scarcely attempted. Some studies~\cite{qi2023fine, li2024images} show that custom fine-tuning of MLLMs on seemingly harmless datasets would compromise their safety alignment. Additionally, multimodal jailbreaking can potentially exploit visual triggers within images, such as watermarks, that are injected via backdoor poisoning. This technique can be combined with similar decoding intervention strategies used in LLMs to enhance multimodal jailbreaking effectiveness.}

\subsection{Limitations and Future Directions on Multimodal Attacks}
While unimodal attacks are extensively studied, multimodal attacks remain underexplored, focusing primarily on textual prompts and image noise with limited exploration in operating multimodal inputs. 

% \noindent\zhuohan{\textbf{Underuse of Model Capabilities} The attacked models' capabilities are not fully leveraged, relying solely on image and text inputs to exploit their understanding of vision and language.}

\noindent\textbf{Unexplored Complex Multimodal Tasks.} Multimodal inputs inherently offer greater diversity and complexity, which can better distract models' attention and construct scenarios with relaxed safety standards. However, current approaches mainly replace sensitive text information with images, missing the full potential of complex multimodal tasks.
% still mimic unimodal strategies by using similar behaviour restriction instructions.
% \noindent\zhuohan{\textbf{Unexplored Image Role} Images are typically used to replace sensitive information in text instructions, such as an object or an event. However, images can take on a more diverse and complex role in instructions, making the input more varied and intricate.}

\noindent\textbf{Neglected Image Domain Shift.} Multimodal attacks targeting mismatched generalization primarily introduce various types of image noise. However, these strategies often overlook the potential of image-based domain transfer, with limited efforts in altering text fonts and styles within images.

% \noindent\textbf{Lack of Harmful Multimodal Training Instances.} There is a notable absence of harmful training instances based on multimodal inputs to disrupt safety alignment. This gap highlights a critical area for future research to develop more sophisticated multimodal training techniques that challenge existing safety mechanisms.

\noindent\textbf{Lack of Multimodal Training Interference.} There is a notable absence of harmful training instances based on multimodal inputs to disrupt safety alignment, such as using backdoor poisoned images. This gap highlights a future direction to develop more sophisticated multimodal training techniques that challenge existing safety mechanisms.

\noindent\textbf{Overly simplistic Attack Generation.} Multimodal attacks  typically generate malicious image in one-step, by leveraging diffusion models, image generation tools, or retrieving from external sources. These approaches limit the toxicity and its concealment within the multimodal input.

% \noindent\zhuohan{\textbf{Simplistic Attack Generation} Multimodal attacks are often generated using straightforward methods, typically by leveraging diffusion models or image generation tools to introduce images.}

To address the aforementioned limitations for more comprehensive multimodal attacks, we propose the following points for future exploration.
\begin{itemize}[itemsep=1pt, leftmargin=13pt, parsep=2pt, topsep=2pt]

\item Explore more diverse multimodal scenarios for context virtualization, where safety standards are more relaxed, such as in science and technology instructional videos. Incorporate more complex multimodal tasks before harmful queries to distract the model's attention, such as complex reasoning games like Jigsaw puzzles.

% \item \zhuohan{Explore diverse roles of images in instructions, such as creating scenarios for context virtualization, acting as standalone tasks like jigsaw puzzles to distract attention, or serving as complete instructions, like a comic strip with text.}

\item Transfer image distribution without altering content by converting to various visual styles (e.g., artistic, animated), adjusting image attributes (scuh as brightness, contrast, saturation), and adding perturbations like mosaic or geometric transformations. Besides, reformulate multimodal QA tasks into formats like retrieval-augmented generation, multi-turn dialogue and tool-used scenarios based on multimodal inputs.

% \item \zhuohan{Transfer image distribution without altering content, for instance, by converting images to different styles like artistic or animated, adjusting attributes like brightness and saturation, adding perturbations like mosaic or geometric transformations, or using specific image types like captchas.}

\item Construct malicious instances with multimodal inputs to disrupt safety alignment during training, such as injecting visual triggers like watermarks, into images through backdoor poisoning.

\item Devise sophisticated multimodal attacks by using iterative methods to refine inputs with model feedback, or by implementing multi-agent systems to collaboratively generate attacks.

% \item \zhuohan{Leverage the model's full range of abilities to make jailbreak success. These abilities can include retrieval, tool usage, image generation and so on.}

\end{itemize}

% \subsection{Summary}
% Overall, these jailbreak attack methods are crafted either manually, leveraging human prior knowledge, or generated automatically via a target-based optimization process, or via collaboration among LLMs. Through this meticulous process, researchers aim to push the boundaries of language model capabilities while simultaneously highlighting potential risks associated with their deployment in real-world applications, and inspire the potential directions to more effective jailbreak defenses for both unimodal and moultimodal models.  

% Advantages and Disadvantages of different types of attack methods. 

% Potential Directions and attempts, especially existing MLLMs.

\section{Jailbreak Defense}
\label{sec:methods_jailbreak_defense}
Jailbreak defense methods protect models from generating harmful content, falling into two main categories: extrinsic and intrinsic defenses. Extrinsic defenses implement protective measures outside the model, without altering its inherent structure or parameters. Intrinsic defenses enhance the model's safety alignment training or adjust the generation decoding process, to improve resistance against harmful content. \emph{We primarily focus on defense strategies for unimodal models as existing research mainly targets LLMs, with a brief overview of multimodal efforts and a discussion of ongoing limitations and potential research directions.}

\subsection{(Unimodal) Extrinsic Defense}
Extrinsic defenses primarily focus on providing pre-safeguard or post-remediation against attacks via plug-in modules or textual prompts. 

\vspace{2mm}
\noindent\textbf{Pre-Safeguard}
There are two strategies for pre-safeguard: harmfulness detection and exposure. 
\begin{enumerate}[itemsep=1pt, leftmargin=13pt, parsep=2pt, topsep=2pt, font=\bfseries]
\item \textbf{Harmfulness Detection.} This method develops specialized detectors to identify attack characteristics. Inspired by the higher perplexity observed in machine-generated adversarial prompts, \citet{alon2023detecting} train a classifier using the Light Gradient-Boosting Machine (LightGBM) algorithm to detect prompts with high perplexity and token sequence length. \citet{kim2023lifetox} fine-tune a RoBERTa-based classifier for implicit toxicity detection across contexts. ~\citet{kumar2023certifying} introduce an erase-and-check framework that individually erases tokens and uses Llama-2~\cite{touvron2023llama2} or DistilBERT~\cite{sanh2019distilbert} to inspect the toxicity of the subsequences, labeling a prompt as harmful if any subsequence is toxic.

\item \textbf{Harmfulness Exposure.} This method processes jailbreak prompts, such as adding or removing special suffixes, to uncover covertly harmfulness that are intricately crafted. 
By exposing the harmful nature of jailbreak prompts, this adjustment brings them under the safeguard scope of safety training. Techniques like smoothing~\cite{robey2023smoothllm, ji2024defending} reduce noise within adversarial prompts through non-semantic-altering perturbations at the character, sentence and structure levels. Translation-based strategies, such as multi-lingual and iterative translation~\cite{yung2024round}, and back-translation~\cite{wang2024defending}, recover the original intent of disguised jailbreak prompts. Additionally, \citet{zhou2024robust} add defensive suffixes or trigger tokens to adversarial prompts through gradient-based token optimization to enforces harmless outputs. 
\end{enumerate}

% Perplexity-reduced?
\vspace{1.5mm}
\noindent\textbf{Post-Remediation}
Unlike pre-safeguard measures, post-remediation allows models to generate responses first, and then modify them to ensure their benignity. For example, \citet{helbling2023llm} prompt LLMs to self-defense by detecting and filtering out potentially harmful content they generate. ~\cite{robey2023smoothllm, ji2024defending} use an ensemble strategy, aggregating predictions from multiple smoothing copies to achieve harmless outputs. A self-refinement mechanism prompts LLMs to iteratively refine their response based on self-feedback to minimize harmfulness~\cite{kim2024break}.

\subsection{(Unimodal) Intrinsic Defense}
There are two main streams to intervene in models' internal training or decoding processes for defense.

\vspace{1.5mm}
\noindent\textbf{Safety Alignment} Improving the safety alignment of large-scale models enhances their robustness against jailbreak attacks, can be achieved by supervised instruction tuning and RLHF. \citet{qi2023fine} implement a simple defense method by incorporating safety examples in the fine-tuning dataset. \citet{bhardwaj2023red} propose red-instruct for safety alignment  by minimizing the negative log-likelihood of helpful responses while penalizing harmful ones. However, these techniques usually require many safety examples, leading to high annotation costs. To address this, \citet{wang2024mitigating} offer a cost-effective strategy using prefixed safety examples with a secret prompt acting as a ``backdoor trigger''. 
\citet{ouyang2022training} adopt RLHF on LLMs to align their behaviour with human preferences, improving performance and safety across various tasks. \citet{bai2022constitutional} replace human feedback with AI feedback, training a harmless but non-evasive AI assistant that responds to harmful queries by constructively explaining its objections. 

\vspace{1.5mm}
\noindent\textbf{Decoding Guidance}
Without tuning the target model, \citet{li2023rain} utilize a Monte-Carlo Tree Searching (MCTS)-style algorithm. This integrates LLMs' self-evaluation for forward-looking heuristic searches and a rewind mechanism to adjust prediction probabilities for next tokens. \cite{xu2024safedecoding} train a safer expert model, and ensemble the decoding probabilities of both the expert model and the target model on several initial tokens, thus enhancing the overall safety of the decoding process.

\subsection{Multimodal Jailbreak Defense}
\emph{Compared to unimodal jailbreak defense, multimodal methods are less explored. An attempt involves translating input images into text and feeding them into LLMs for safer response, using unimodal pre-safeguard strategies~\cite{gou2024eyes}. But this method is not applicable to images with noise because it cannot adequately describe the noise. 
% Given that attack image queries usually contain complex perturbations that reduce response robustness, 
To address complex perturbations in attack images,
\citet{zhang2023mutation} propose to mutate inputs into variant queries and check for response divergence to detect jailbreak attacks. \citet{zong2024safety} advance multimodal safety alignment by constructing an instruction-following dataset, VLGuard, for safety fine-tuning of MLLMs.}

\subsection{Limitations and Future Directions on Multimodal Defense}
While unimodal defense methods still need improvement, the less-explored multimodal defenses require further research with limitations as follows:
% . Overall, multimodal defense methods have the following limitations.

\noindent \textbf{Non-generalizable Defense.}
Most defense strategies are tailored to specific attack types, struggling to adapt to various and evolving attack methods.
% There is an urgent need for a comprehensive and evolvable defense system that can adapt to continuously evolving attack techniques.

\noindent \textbf{Poor Robustness.} Existing defenses struggle to withstand perturbation attacks, where subtle and imperceptible changes to inputs can cause failures in detecting jailbroken content. Developing robust defenses against attacks is a significant challenge.

\noindent \textbf{False Positive Challenge.} Legitimate responses may be excessively defended and wrongly flagged as jailbreak attacks, hindering user needs.

\noindent\textbf{High Cost of Safety Alignment.}
Fine-tuning for safety requires extensive annotation, leading to high costs. Besides, repeated alignment training due to models advancements and evolving attack methods, incurs high computation expenses.

% \noindent\textbf{\zhuohan{Unexplored Techniques for Image Defense}.} Current methods for detecting harmful content in images primarily rely on classifying textual descriptions. Direct image-based detection and smoothing techniques still need further research.

\noindent\textbf{Unexplored Image-based Detection.} Current methods primarily detecting harmful content in images based on their textual descriptions. Direct detection and smoothing techniques that operate on images still need further research.

To address these challenges, we propose the following research directions:
\begin{itemize}[itemsep=1pt, leftmargin=13pt, parsep=2pt, topsep=2pt]
\item Develop a comprehensive and adaptable defense system for evolving attack techniques. For example, ensemble multiple defense strategies at various stages, or design a general reinforcement learning algorithm to optimize strategies through simulated attack-defense scenarios.
\item Regularly update adversarial training sets with new examples from recent attack trends and continuously train a defense model, to improve resilience against perturbation-based attacks.
\item Design fine-grained defense methods to identify varying degrees of harmfulness, and adjust thresholds accordingly in different scenarios. Besides, utilize majority-vote or cross-validation to mitigate false positive issues.
\item Identify subsets within fine-tuning datasets that, although benign, may degrade model safety and remove them for subsequent tuning. Besides, implement model pruning to update specific sub-regions for safety alignment.
\item Explore detection and smoothing techniques that directly classify and mitigate harmful content in images inputs.
\end{itemize}

\section{Conclusion}
\label{sec:conclusion}
In this work, we offer a thorough overview of jailbreaking research for LLMs and MLLMs, discussing recent advances in evaluation benchmarks, attack techniques and defense strategies. Furthermore, we summarize the limitations and potential research directions of of MLLM jailbreaking by drawing comparisons to the more advanced state of LLM jailbreaking, aiming to inspire future work.

\section*{Limitations}
This study has several potential limitations. First, due to space constraints, we may not include all relevant references and detailed technical methods related to jailbreaking. 
Second, our work is primarily focused on highlighting limitations and potential research directions in the multimodal domain, while not providing an in-depth analysis of unimodal limitations. Finally, this work mainly serves as a survey and investigation on existing and future jailbreak research, without proposing and experimenting with specific novel methods.

\section*{Ethics Statement}
This paper discusses jailbreak datasets and attack techniques, which may potential contain or induce offensive and harmful content. It is important to emphasize that this work aims to inspire future research on jailbreaking to enhance the robustness and security of large models, aiding in the identification and mitigation of potential vulnerabilities. We strongly urge more researchers to focus on this area to promote the development of more ethical and secure large models. Our survey and discussed content are strictly intended for research purposes that follow the ethical guidelines of the community. The authors emphatically denounce the use of our work for generating harmful content.

\bibliography{custom}

\appendix

\section{Evaluation Framework}
\label{sec:appendix}
The evaluation of jailbreak attack and defense involves three key factors.
First, the definition of a successful jailbreak builds a standard for response assessment. Second, the metrics which quantitatively measure the effectiveness of specific attack or defense strategies. The third is the judgement methods, which aim to accurately assess results and align with human values. Subsequent paragraphs will detail existing research to these points.

\paragraph{Definitions of Successful Jailbreak}
A successful jailbreak attack can be determined at three different levels.
The most basic level deems an attack successful if the response does not directly reject the query (i.e., lacks words related to rejection)~\cite{zou2023universal,robey2023smoothllm}. This conservative approach is only appropriate for scenarios demanding explicit rejection. However, in most contexts, a more suitable response aligning with human values might be a well-rounded statement or an ethical recommendation~\cite{wang2023not}. A more applicable criterion considers an attack successful if the model produces on-topic and harmful responses~\cite{wei2024jailbroken,yong2023low,yu2023gptfuzzer,wang2023not,deng2024masterkey,attacksloft,zhan2023removing,shah2023scalable}, focusing on whether output content circumvent safety mechanisms without assessing the response quality, like its potential harm or benefit to the attacker.
The most stringent definition assesses both the content and the impact of responses, identifying an attack as successful if it contains substantially harmful content and aids harmful actions~\cite{huang2023catastrophic,chao2023jailbreaking,souly2024strongreject,ji2024defending}.

\paragraph{Evaluation Metrics}
The evaluation of jailbreak primarily utilizes two types of metrics: ratio-based and score-based. Ratio-based metrics assess individual responses as a binary classification of a success or failure, calculating an overall rate, such as the attack success rate (ASR)~\cite{wei2024jailbroken,yong2023low,liu2023autodan,robey2023smoothllm,xu2023cognitive,deng2024masterkey,yuan2023gpt,attacksloft,shah2023scalable}. Some studies further distinguishing responses based on compliance levels~\cite{yu2023gptfuzzer} or categories~\cite{wang2023not}, which are then aggregated into an overall success or failure rate. Score-based metrics assign continuous scores to responses, providing a more fine-grained assessment. These scores evaluate aspects like specificity, persuasiveness~\cite{souly2024strongreject,ji2024defending}, detail~\cite{chao2023jailbreaking}, or harmfulness~\cite{huang2023catastrophic}, averaging across the dataset for a comprehensive evaluation.

\paragraph{Jailbreaking Judgement Methods}
Jailbreak attempt assessments utilize various methods.
Human evaluation involves experts manually reviewing responses based on predefined guidelines, ensuring accuracy but at the cost of time and scalability~\cite{wei2024jailbroken,yong2023low,wang2023not,liu2023jailbreaking,attacksloft,zhan2023removing}.
Rule-based evaluation employ criteria like sub-string matching for rejection keywords, offering cost-effectiveness and ease of implementation, yet lacking flexibility for diverse scenarios and often incompatible with new models due to varying rejection keywords~\cite{zou2023universal,liu2023autodan,robey2023smoothllm,xu2023cognitive}.
Structuring queries for limited response formats, like yes/no~\cite{wang2023decodingtrust} or multiple-choice questions~\cite{xu2023cvalues}, simplifies evaluation but doesn't fully reflect real-world performance, creating a gap in effectiveness.

Model-based evaluation including utilizing official APIs like Perspective API for detecting harmful content~\cite{wang2023decodingtrust}, prompting LLMs as evaluators~\cite{wang2023not,souly2024strongreject,chao2023jailbreaking,yuan2023gpt,shah2023scalable,liu2023autodan}, and training PLM-based evaluators with annotated data~\cite{yu2023gptfuzzer,wang2023not,huang2023catastrophic}. These approaches balance efficiency and flexibility, and aligning well with human values. However, it presents several limitations: LLM-based evaluators are costly and can yield high false-negative rates~\cite{shah2023scalable}, while PLM-based evaluators require extensive human-annotated training data and may suffer from lower accuracy due to imbalanced data distribution~\cite{wang2023not}.

\end{document}